\documentclass[12pt]{article}

\usepackage{amssymb,amsmath,amsfonts,latexsym,graphicx}
\usepackage[numbers]{natbib}
\usepackage{multicol}
\usepackage{caption}
\usepackage{float}

\usepackage{wrapfig}        
\usepackage{enumitem}       

\usepackage[format=plain]{caption}
\usepackage{subcaption}

\usepackage{booktabs}
\usepackage{multirow}
\usepackage[colorlinks=true, linkcolor=blue, urlcolor=blue, citecolor=blue]{hyperref}
\usepackage{tabularx}

\begin{document}
\begin{center}

\Large \bf Domain-Adaptive Pre-training of Self-Supervised Foundation Models for Medical Image Classification in Gastrointestinal Endoscopy\rm

\vspace{1cm}


\large Marcel Roth$\,^1$\textsuperscript{*†}, \large Micha V. Nowak$\,^1$, Adrian Krenzer$\,^1$, Frank Puppe$\,^1$

\vspace{0.5cm}

\normalsize


$^1$ Artificial Intelligence and Knowledge Systems,\\
Julius-Maximilians-Universität Würzburg, Sanderring 2, Würzburg,
97070, Germany.


\vspace{5mm}

\textsuperscript{*}Corresponding author: \texttt{\href{mailto:marcel.roth@stud-mail.uni-wuerzburg.de}{marcel.roth@stud-mail.uni-wuerzburg.de}} \\
E-mail(s): \texttt{\href{mailto:micha.nowak@stud-mail.uni-wuerzburg.de}{micha.nowak@stud-mail.uni-wuerzburg.de}}
\texttt{\href{mailto:adrian.krenzer@uni-wuerzburg.de}{adrian.krenzer@uni-wuerzburg.de}}\\
\texttt{\href{mailto:frank.puppe@uni-wuerzburg.de}{frank.puppe@uni-wuerzburg.de}}
 \\
\textsuperscript{†}Primary contributor

\end{center}

\abstract{
Video capsule endoscopy has transformed gastrointestinal endoscopy (GIE) diagnostics by offering a non-invasive method for capturing detailed images of the gastrointestinal tract, enabling early disease detection. However, its potential is limited by the sheer volume of images generated during the imaging procedure, which can take anywhere from 6-8 hours and often produce up to 1 million images, necessitating automated analysis. Additionally, the variability of these images, combined with the need for expert annotations and the scarcity of large, high-quality labeled datasets, constrains the effectiveness of current medical image analysis models. To address this, we introduce a novel large GIE dataset, called EndoExtend24, created by merging ten existing public and private datasets, ensuring patient integrity across splits. EndoExtend24 includes over 226,000 labeled images, as well as dynamic class mappings, which allow unified training across datasets with differing labeling granularity, supporting up to 123 distinct pathological findings. Further, we propose to leverage domain adaptive pre-training of foundation models trained with self-supervision on generic image data, to adapt them to the task of GIE medical image diagnosis. Specifically, the EVA-02 model, which is based on the ViT architecture and trained on ImageNet-22k with masked image modeling (using EVA-CLIP as a MIM teacher), is pre-trained on the EndoExtend24 dataset to achieve domain adaptation, and finally trained on the Capsule Endoscopy 2024 Challenge dataset.

Our model demonstrates robust performance, securing third place in the Capsule Endoscopy 2024 Challenge. We achieved a macro AUC of 0.762 and a balanced accuracy of 37.1\% on the test set, significantly outperforming the best baseline model, ResNet50V2, which achieved a macro AUC of 0.542 and a balanced accuracy of 17.7\%. Notably, we surpassed the first-place model in balanced accuracy, with our 37.1\% compared to their 35.7\%, despite their higher macro AUC of 0.857. These results emphasize the effectiveness of our domain-adaptive pre-training approach and the enriched EndoExtend24 dataset in advancing gastrointestinal endoscopy diagnostics.
}

\newpage
\section{Introduction}\label{sec1}
Video capsule endoscopy (VCE) has revolutionized gastrointestinal endoscopy (GIE) diagnostics by providing a non-invasive way to capture high-resolution images of the GI tract, facilitating early detection of various diseases. Additionally, it significantly improves accessibility, as it can be deployed in areas with limited medical infrastructure and enables remote diagnostics.
However, a key limitation of this technology is the massive volume of data generated -- often exceeding 1 million frames over the course of a 6-8 hour procedure -- making manual analysis infeasible and necessitating the development of automated solutions. Current medical image analysis models face challenges due to the variability in image quality, the requirement for expert annotations, and the lack of large, high-quality labeled datasets. To address these challenges, we introduce EndoExtend24, a novel, large-scale gastrointestinal endoscopy dataset comprising over 226,000 labeled images. This dataset is constructed by merging and re-stratifying the train/test splits of ten existing public and private datasets, ensuring no patient data overlap between splits. The dataset includes dynamic class mappings that accommodate variations in labeling granularity across the original datasets, supporting up to 123 distinct pathological findings.

\begin{figure}[H]
    \captionsetup[subfigure]{labelformat=empty, skip=2pt}
    \newcommand{\verysmall}{\fontsize{5pt}{8pt}\selectfont}
    \centering
    
    \begin{minipage}{\textwidth}
    \captionof{table}[Multi-class GIE datasets that are part of EndoExtend24.]{\textbf{Multi-class GIE datasets that are part of EndoExtend.} Datasets are grouped by their split type and contain images obtained via Video Capsule Endoscopy (VCE), Gastroscopy (GST), and Colonoscopy (COL). \textsuperscript{*}The SEE-AI Project dataset employed a 3-fold CV split in the paper, but did not provide this split publically.}
    \label{tab:dataset_framework:method:multi_class_datasets}
    \setlength{\tabcolsep}{6pt}
    \centering
    \footnotesize
    \begin{tabularx}{\textwidth}{llXclrr}
    \toprule
    \textbf{Split Type} & \textbf{Year} & \textbf{Dataset} & \textbf{Modality} & \textbf{Resolution} &  \textbf{Classes} & \textbf{Images} \\
    \midrule
    \multirow{3}{*}{None} 
         & 2023 & The SEE-AI Project\textsuperscript{*} \cite{SEEAIDataset} & VCE & 576×576 & 12 & 18,481 \\ 
         & 2017 & KID 2 \cite{KID1_2_Atlas, KID2} & VCE & 360×360 & 8 & 2,371\\ 
         & 2017 & KID 1 \cite{KID1_2_Atlas, KID1_Paper1, KID1_Paper2} & VCE & 360×360 & 9 & 77 \\ 
    \cmidrule{2-7}
    \multirow{3}{*}{Patient-ID} 
         & 2022 & ERS \cite{ERS2022} & GST, COL, VCE& various & 123 & 121,399 \\ 
         & 2022 & LIMUC \cite{LIMUC2022} & COL & 352×288 & 4 & 11,276 \\ 
         & 2023 & MedFMC \cite{MedFMC2023} & COL & 1280×1024 & 4 & 3,865 \\ 
    \cmidrule{2-7}
    \multirow{3}{*}{K-Fold CV} 
         & 2021 & Kvasir-Capsule \cite{KvasirCapsule2021} & VCE & 336×336 & 14 & 47,238 \\ 
         & 2020 & HyperKvasir \cite{HyperKvasir2020} & GST, COL & various & 23 & 10,662 \\ 
         & 2020 & CrohnIPI \cite{CrohnIPI} & VCE & 320×320 & 7 & 3,498 \\ 
    \cmidrule{2-7}
    \multirow{1}{*}{Train-Val-Test} 
         & 2023 & GastroVision \cite{GastroVision2023} & GST, COL & various & 27 & 8,000 \\ 
    \midrule[1pt]
    & 2024 & \textbf{EndoExtend (Ours)} &&&& \textbf{226,867} \\
    \bottomrule
    \end{tabularx}
    \end{minipage}

    \vspace{0.5cm}
    
    \begin{subfigure}[b]{0.11\textwidth}
        \centering
        \includegraphics[width=\textwidth]{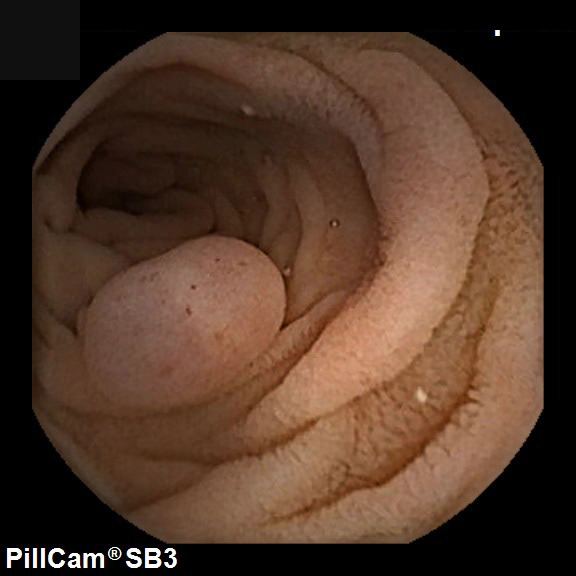}
        \caption{\verysmall SEE-AI}
    \end{subfigure}
    \hspace{-7pt}
    \begin{subfigure}[b]{0.11\textwidth}
        \centering
        \includegraphics[width=\textwidth]{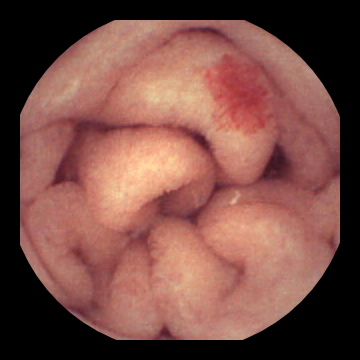}
        \caption{\verysmall KID}
    \end{subfigure}
    \hspace{-7pt}
    \begin{subfigure}[b]{0.11\textwidth}
        \centering
        \includegraphics[width=\textwidth]{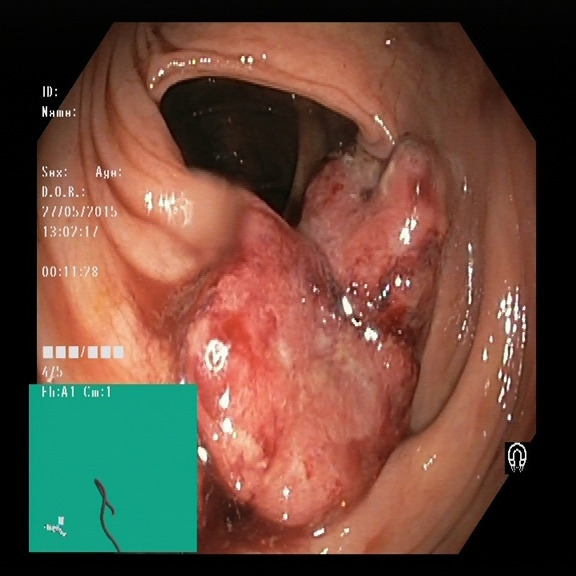}
        \caption{\verysmall ERS}
    \end{subfigure}
    \hspace{-7pt}
    \begin{subfigure}[b]{0.11\textwidth}
        \centering
        \includegraphics[width=\textwidth]{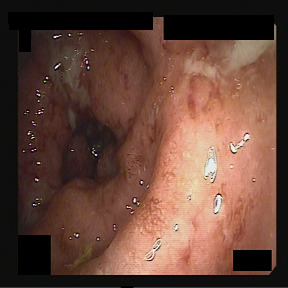}
        \caption{\verysmall LIMUC}
    \end{subfigure}
    \hspace{-7pt}
    \begin{subfigure}[b]{0.11\textwidth}
        \centering
        \includegraphics[width=\textwidth]{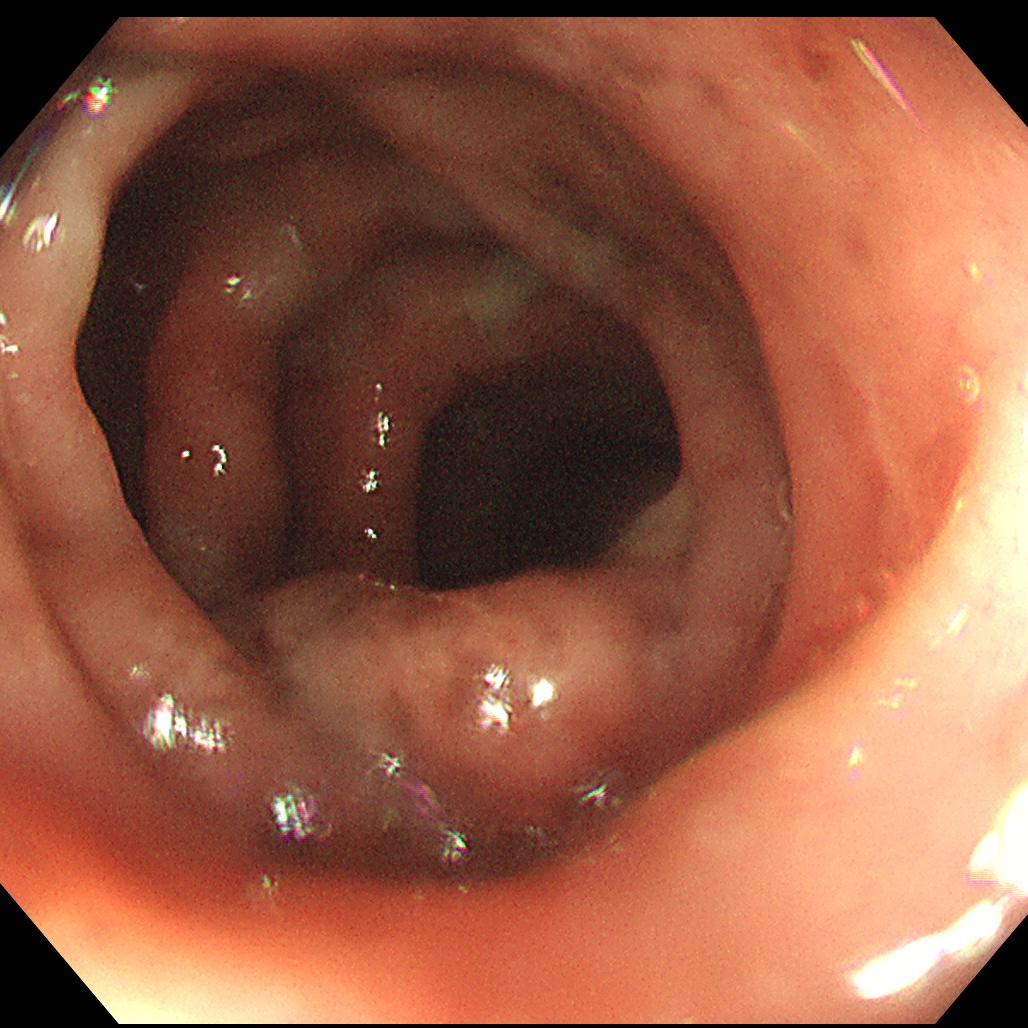}
        \caption{\verysmall MedFMC}
    \end{subfigure}
    \hspace{-7pt}
    \begin{subfigure}[b]{0.11\textwidth}
        \centering
        \includegraphics[width=\textwidth]{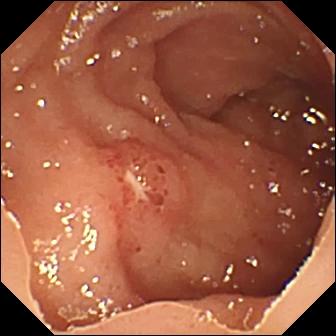}
        \caption{\verysmall Kvasir-Capsule}
    \end{subfigure}
    \hspace{-7pt}
    \begin{subfigure}[b]{0.11\textwidth}
        \centering
        \includegraphics[width=\textwidth]{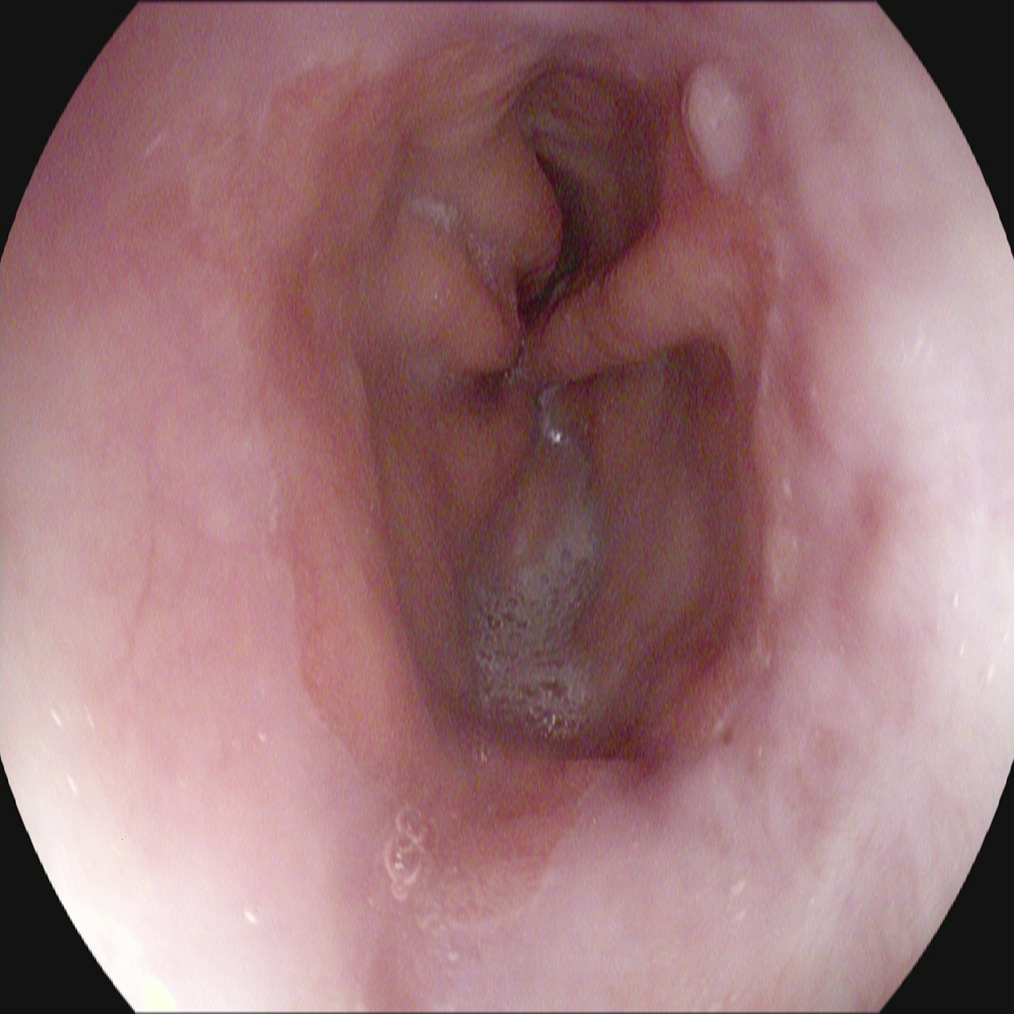}
        \caption{\verysmall Hyper-Kvasir}
    \end{subfigure}
    \hspace{-7pt}
    \begin{subfigure}[b]{0.11\textwidth}
        \centering
        \includegraphics[width=\textwidth]{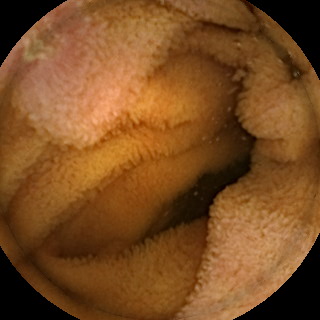}
        \caption{\verysmall CrohnIPI}
    \end{subfigure}
    \hspace{-7pt}
    \begin{subfigure}[b]{0.11\textwidth}
        \centering
        \includegraphics[width=\textwidth]{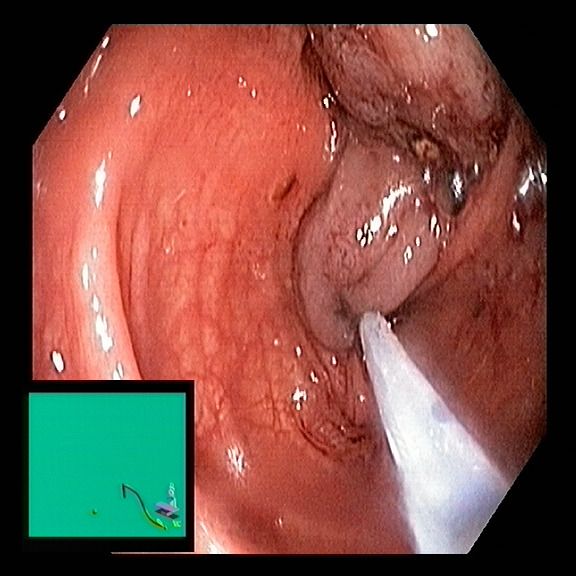}
        \caption{\verysmall GastroVision}
    \end{subfigure}

    \captionof{figure}{Sample frames obtained from the datasets included in EndoExtend24.}
    \label{fig:method:dataset_images}
\end{figure}

We further propose leveraging the EVA-02 model, a vision transformer pretrained on ImageNet-22k with masked image modeling (MIM) using EVA-CLIP as a teacher model, for domain adaptation in GIE diagnostics. The EVA-02 model incorporates advanced architectural features like SwiGLU activations, Rotary Position Embeddings (ROPE), and additional Layer Normalization (LN) in its MLP layers. Pretraining the EVA-02 model on the EndoExtend24 dataset allows it to learn domain-specific features, before being fine-tuned on the Capsule Endoscopy 2024 (CE24) Challenge dataset. Our experimental results demonstrate the effectiveness of this approach, achieving an AUC Macro score of 0.993 and a balanced accuracy of 89.3\% on the challenge validation set.

\section{EndoExtend24}\label{sec2}
In the field of gastrointestinal endoscopy, acquiring a large and diverse collection of high-quality labeled images presents a complex task for driving deep learning (DL) applications, characterized by various fundamental challenges:

\begin{enumerate}[label=(\arabic*)]

 \item  \textbf{Dataset Shortage \& Accessibility:} Rare disease
occurrences, coupled with stringent requirements for expert annotations by trained
medical professionals are crucial for robust  model training and create significant bottlenecks. The medical domain’s complexity is compounded by ethical considerations and privacy concerns, making the
acquisition of sufficient quantities of annotated data particularly difficult. The shortage and limited accessibility of multi-class GIE datasets represents a significant challenge for researchers, often necessitating the merge of multiple datasets to obtain an
appropriate quantity of data for effective machine learning training.

\item \textbf{Terminology and Standardization:} The integration of diverse datasets becomes a challenging and error-prone task when a universally accepted terminology and standard is absent. In the field of GIE, the Minimal Standard Terminology (MST) \cite{MST3.0} represents a significant effort to standardize the terminology used, developed by the European Society of Gastrointestinal Endoscopy to ensure clear and consistent reporting of endoscopic findings. This standardized terminology provides a comprehensive set of definitions, descriptions, and classifications, allowing both medical researchers and computer science researchers to develope machine learning (ML) solutions that accurately describe diagnostic and therapeutic procedures. By establishing a common language, MST enhances the quality of clinical data, supports study comparability, and improves patient care. However, the lack of consistency in terminology across different publicly available datasets, with only the ERS dataset \cite{ERS2022} fully conforming to the MST standard, poses significant challenges. Disparate naming conventions, terminology, granularity, and class mappings across datasets introduce data inconsistencies that complicate data collection, pre-processing, analysis, and integration. These discrepancies ultimately diminish the reliability of model outcomes, hinder research progress, and adversely affect the comparability and trustworthiness of scientific studies, impacting the quality of the utilized ML techniques.

 \item \textbf{Data Leakage:} The widespread absence of patient ids (PIDs) among datasets complicates the process of dividing the dataset into distinct training, validation, and test sets, potentially leading to data leakage and an overestimation of model performance \cite{VanDerSommen2020MachineField}. Data leakage occurs when information from the same patient --- typically linked to a PID --- inadvertently appears in multiple splits, such as both the training and validation sets, resulting in misleadingly high performance metrics. This is a frequent issue in medical and especially in GIE datasets, which is underscored by EndoExtend24 in Table \ref{tab:dataset_framework:method:multi_class_datasets}. Only 30\% (3 out of 10) of the used datasets provide patient information while another 30\% provide no split meta information at all.

 \item \textbf{Split Flexibility:} The lack of flexibility in dataset splits presents yet another significant issue. Most GIE datasets offer either a single predefined train-validation-test split, a K-Fold CV split, or provide neither patient information nor split details. As a general guideline for the provision of a dataset in a medical context, it is always highly preferable to include available information regarding the split of a dataset rather than no information at all. Dealing with predefined split information can prove challenging when a specific test ratio and amount of CV folds is desired, thereby further complicating the dataset splitting process.

\end{enumerate}

To address the challenges outlined above, we introduce EndoExtend24, a large-scale, multi-class GIE dataset comprising over 226,000 labeled images sourced from ten public and private datasets. EndoExtend24 is designed to support DL research in GIE by overcoming issues related to data scarcity, inconsistent standards, data leakage and split flexibility. As shown in Table \ref{tab:dataset_framework:method:multi_class_datasets}, the dataset incorporates three examination modalities: video capsule endoscopy, gastroscopy, and colonoscopy. Furthermore, it offers a highly flexible class-mapping structure, encompassing up to 123 fine-grained pathological findings across diverse imaging resolutions. This extensive variety allows researchers to explore more granular classifications while maintaining interoperability between datasets. The split types vary from datasets that offer no predefined split metadata, such as the SEE-AI Project, to those providing essential PID information, K-Fold cross-validation (CV) splits, or standard train-validation-test splits.

Furthermore, along with EndoExtend24, we introduce a framework that allows the seamless integration of new datasets via dataset adapters, promoting dataset interoperability. Within the framework, we implement and provide a stratified group splitting algorithm, as well as a stratified k-fold cross validation splitting algorithm. The novelty lies in the framework both respecting the patient integrity while maintaining class stratification across splits.

\section{Methodology}\label{sec3}

\subsection{Dataset Preparation}\label{sec3:subsec1}

\paragraph{Data Integrity}
The majority of the images within the KID, SEE-AI, and KvasirCapsule datasets is part of both the EndoExtend24 and the CE24 datasets. Since CE24 applies cropping as a pre-processing step, the images in EndoExtend24 and CE24 are not identical. Moreover, CE24 does not contain the entire datasets. To prevent data leakage -- which leads to an overestimation of validation performance -- it is crucial to maintain strict separation between training and validation data throughout the entire model training process. We address this as follows:

\begin{enumerate}[label=(\arabic*)]
    \item First, we increase the CE24 training set from 70\% to 80\%, transferring 10\% of the validation dataset to the train dataset while preserving class stratification, resulting in a 20\% validation split.
    \item Next, we exclude all overlapping datasets (KID, SEE-AI, and KvasirCapsule) from the splitting process and split the remainder of EndoExtend24 likewise into 80\% training and 20\% validation.
    \item Finally, we utilize the CE24 training and validation split (without AIIMS) and enforce it on the original version of the overlapping datasets. The data matching is carried out using the filenames, as these remained unaltered.
\end{enumerate}

\begin{figure}[H]
    \centering
    \includegraphics[width=1\textwidth]{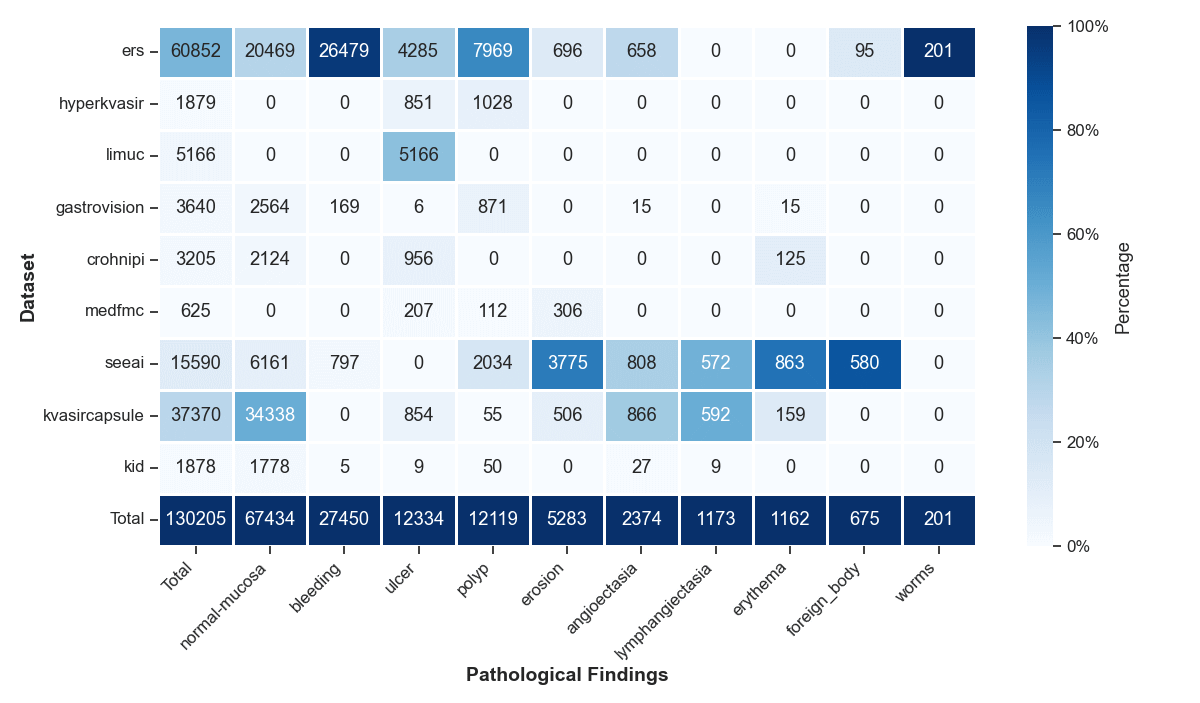}
    \caption{Distribution of pathological findings across the different datasets included in \\EndoExtend24 which was utilized during pre-training.}
    \label{fig:method:dataset_class_dist}
\end{figure}

\paragraph{Subset Selection}
The subset selection of the EndoExtend24 dataset employed for the pre-training in this work was selected to encompass the same 10 pathological findings as those covered in the CE24 dataset, namely: \textit{normal mucosa}, \textit{bleeding}, \textit{ulcer}, \textit{polyp}, \textit{erosion}, \textit{angioectasia}, \textit{lymphangiectasia}, \textit{erythema}, \textit{foreign body}, and \textit{worms}. We selected the same classes for the EndoExtend24 dataset.
The class distribution across the various datasets is shown in Figure \ref{fig:method:dataset_class_dist}. 
Each cell in the heatmap represents the sample count for a specific pathological finding (class) in a given dataset, with the color intensity indicating the percentage that dataset contributes to a specific class. There are notable differences in sample distribution between datasets. For instance, ERS includes 20,469 samples of normal mucosa, while KvasirCapsule has 34,338. 

LIMUC, on the other hand, focuses primarily on ulcers, with 5,166 samples, whereas HyperKvasir provides fewer samples, such as 1,028 for polyps and 851 for ulcers. 
The combined sample counts for each finding are summarized in the bottom row of the heatmap. 
Normal mucosa is the most common with 67,434 samples, followed by bleeding with 27,450, and ulcer with 12,334. Some findings are comparatively underrepresented, such as foreign bodies (675 samples) and worms (201 samples). 
These imbalances, common in medical datasets, require careful consideration during dataset splitting and the use of techniques such as weighted random sampling during training to ensure balanced representation of all categories.

\subsection{Train Augmentations}
To enhance the generalization of our model, we applied a series of data augmentations, both during pre-training and downstream task training, using the \texttt{albumentations} library. The augmentations include spatial transformations such as horizontal and vertical flips, random 90-degree rotations, and grid distortions, which help simulate various viewing angles and deformations typical in video capsule endoscopy images. We also used Gaussian blur with a variable kernel size and applied random resized cropping to introduce variability in scale. Color-based augmentations, including color jitter (adjusting brightness, contrast, saturation, and hue), further increased the diversity of the training data by simulating different lighting conditions. Finally, images were normalized using mean and standard deviation values based on the ImageNet dataset and converted to tensor format for compatibility with \texttt{PyTorch}. For validation, we employed a simpler augmentation pipeline, using only resizing, center cropping, normalization, and tensor conversion to maintain consistency and avoid altering the image content. Both training and validation images were resized to 224$\times$224 pixels.

\subsection{Pretraining}\label{sec3:subsec2}
For pre-training, the model checkpoint \texttt{timm/eva02\_base\_patch14\_224.mim\_in22k} was obtained via the HuggingFace Model Hub. This specific checkpoint was chosen due to its strong baseline performance on various computer vision tasks, providing a solid foundation for domain adaptation. We utilized the EndoExtend24 dataset, previously described, to perform domain-adaptive pre-training. The goal of this phase was to align the general pre-trained EVA-02 model to the specific domain of GIE. To achieve this, we trained the model using a learning rate of 1e-6 over 50 epochs with a batch size of 64. The AdaBelief optimizer was employed to manage updates efficiently \cite{adabelief}. Domain adaptation was crucial in this phase to enhance the model’s ability to extract relevant features in GIE, which differs substantially from the original pre-training domain. We also evaluated various architectures alongside EVA-02, such as SEER \cite{SEER}, a cross-domain model designed for robust transfer learning across domains. After performance evaluations on a validation subset, EVA-02 emerged as the best-performing model in terms of generalization and transferability, and it was selected for subsequent downstream training on task-specific data.

\subsection{Model Training}\label{sec3:subsec3}

Following the domain adaptation phase, the pre-trained model underwent transfer learning on the CE24 training dataset. While the previous step adapted the model to the broader domain of GIE, this final training process was specifically focused on narrowing the domain further to VCE. The CE24 dataset consists of a curated collection of capsule endoscopic imagery, which represents the specific data format and clinical context the model would encounter in practical use, such as lesion detection and classification in capsule-based gastrointestinal imaging. To train the model, we used a learning rate of 1e-6, dynamically adjusted using a lambda scheduler. The learning rate was reduced by a factor of 0.95 after each epoch, ensuring a smooth and controlled convergence. The AdaBelief optimizer was chosen for this stage due to its ability to provide fast convergence while maintaining strong generalization capabilities, especially useful in adapting the model to the specific challenges of capsule endoscopy. Training was conducted over 50 epochs with a batch size of 128, which allowed for efficient learning while ensuring the model could adapt to the nuances of video capsule imagery. Instead of employing early stopping, we manually selected a lower number of epochs to mitigate overfitting based on observed validation performance trends. By closely monitoring training and validation metrics, we ensured that the model progressively aligned with the unique visual and diagnostic characteristics of VCE, refining its ability to perform tasks such as lesion detection and classification in this domain.

\section{Evaluation}
The performance of our final eva02-base model was evaluated on the CE24 validation set, consisting of 10,658 images. Table \ref{tab:results_validation} presents a comparative analysis of our model's performance against baseline models, both after pre-training and on the final downstream task. The eva02-base model demonstrates superior performance across all metrics in both stages.

\begin{table}[H]
\begin{minipage}{\textwidth} 
\centering
\caption{Performance of our final eva02-base model after pre-training on the EndoExtend24 dataset and fine-tuning on the CE24 validation set. Accuracy (ACC) refers to the balanced accuracy, F1, AUC, and mAP are reported as macro average. The evaluation was carried out by ourselves.}
\label{tab:results_validation}
\begin{tabular}{@{}ll|l|l|l|l@{}}
\toprule
  & \textbf{Model Name} & \textbf{ACC} & \textbf{AUC} & \textbf{F1} & \textbf{mAP}\\
  
\midrule 
\multirow{2}{*}{Pre-Training} 
& eva02-base 
& \textbf{0.810}  
& \textbf{0.976}
& \textbf{0.786}
& \textbf{0.860}
\\ 
& SEER
& 0.743 
& 0.960 
& 0.723
& 0.755
\\

\cmidrule{1-6}

\multirow{5}{*}{Downstream} 
& eva02-base \textit{(ours)}
& \textbf{0.893}
& \textbf{0.993}
& \textbf{0.875}
& \textbf{0.931}
\\ 
& VGG16
& 0.568
& 0.916
& 0.484
& 0.525
\\
& SVM
& 0.41 
& 0.94
& 0.49
& N/A
\\
& ResNet50
& 0.320
& 0.871
& 0.370
& N/A
\\
& Custom CNN
& 0.10
& N/A
& 0.09
& N/A
\\

\bottomrule
\end{tabular}
\end{minipage}
\end{table}

In our own evaluation of the downstream task, our model achieved a balanced accuracy of 0.893, a macro AUC of 0.993, a weighted F1 score of 0.960, and a macro mAP of 0.931 on the CE24 validation set. The final test results of the Capsule Video Endoscopy Challenge 2024 are presented in Table \ref{tab:results_test}, where our approach secured third place among 27 participating teams. On the validation set, however, we clearly outperformed all other teams in terms of balanced accuracy, achieving a score of 0.916, and in the combined metric, attaining 0.954. Notably, we achieved a 6.6\% higher combined metric compared to the top-ranked team, which had a score of 0.888. This discrepancy between our superior validation performance and our overall placement suggests that our strategy of optimizing the models for balanced accuracy, rather than for the combination of mean AUC and balanced accuracy used for the final ranking, may have influenced our test results.

\begin{table}[H]
\centering
\caption{Comparison of baseline models and the top three teams on the CE24 validation and test datasets using goal metrics (mean AUC, balanced accuracy), and the combined metric (mean of AUC + ACC). The evaluation was carried out by the challenge organization team.}
\label{tab:results_test}
\begin{tabular}{@{}c l l cc c c cc@{}}
\toprule
\multirow{2}{*}{} & \multicolumn{2}{c}{} & \multicolumn{2}{c}{\textbf{Mean AUC}} & \multicolumn{2}{c}{\textbf{Balanced ACC}} & \multicolumn{2}{c}{\textbf{Combined}} \\
\cmidrule(lr){4-5} \cmidrule(lr){6-7} \cmidrule(lr){8-9}
& \textbf{} & \textbf{Name} & \textbf{Val} & \textbf{Test} & \textbf{Val} & \textbf{Test} & \textbf{Val} & \textbf{Test} \\
\midrule
\multirow{6}{*}{\rotatebox{90}{Baselines}} 
& 1 & VGG19            & 0.856 & 0.525 & 0.624 & 0.144 & 0.740 & 0.335 \\
& 2 & Xception         & 0.751 & 0.534 & 0.487 & 0.131 & 0.619 & 0.333 \\
& 3 & ResNet50V2       & 0.797 & 0.542 & 0.609 & 0.177 & 0.703 & 0.360 \\
& 4 & MobileNetV2      & 0.926 & 0.548 & 0.536 & 0.114 & 0.731 & 0.331 \\
& 5 & InceptionV3      & 0.772 & 0.524 & 0.509 & 0.128 & 0.641 & 0.327 \\
& 6 & InceptionResNetV2& 0.614 & 0.523 & 0.253 & 0.147 & 0.433 & 0.335 \\
\midrule
\multirow{3}{*}{\rotatebox{90}{Teams}} 
& 1 & PuppyOps \cite{PuppyOps}    & 0.991 & \textbf{0.857} & 0.785 & 0.357 & 0.888 & \textbf{0.607} \\
& 2 & MedInfoLab \cite{MedInfoLab} & \textbf{0.994} & 0.774 & 0.846 & \textbf{0.372} & 0.920 & 0.573 \\
& 3 & WueVision \textit{(ours)} & 0.992 & 0.763 & \textbf{0.916} & 0.371 & \textbf{0.954} & 0.567 \\
\bottomrule
\end{tabular}
\end{table}

\begin{figure}[t!]
    \centering
    \includegraphics[width=1\textwidth]{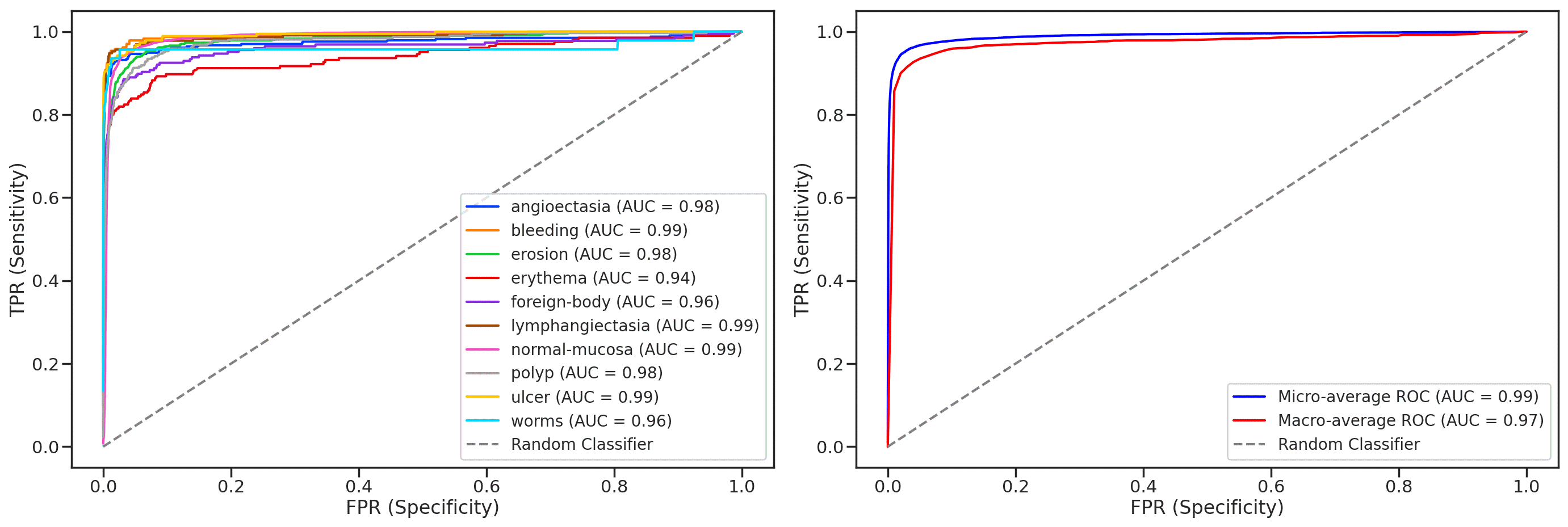}
    \caption{ROC curves of eva02-base on the downstream task of GIE multi-class classification on the CE24 validation dataset. Left: Individual ROC curves for 10 distinct pathological findings, with corresponding AUC values. Right: Micro- and macro-average ROC curves.}
    \label{fig:roc_curves}
\end{figure}

\begin{figure}[b!]
    \centering
    \includegraphics[width=.73\textwidth]{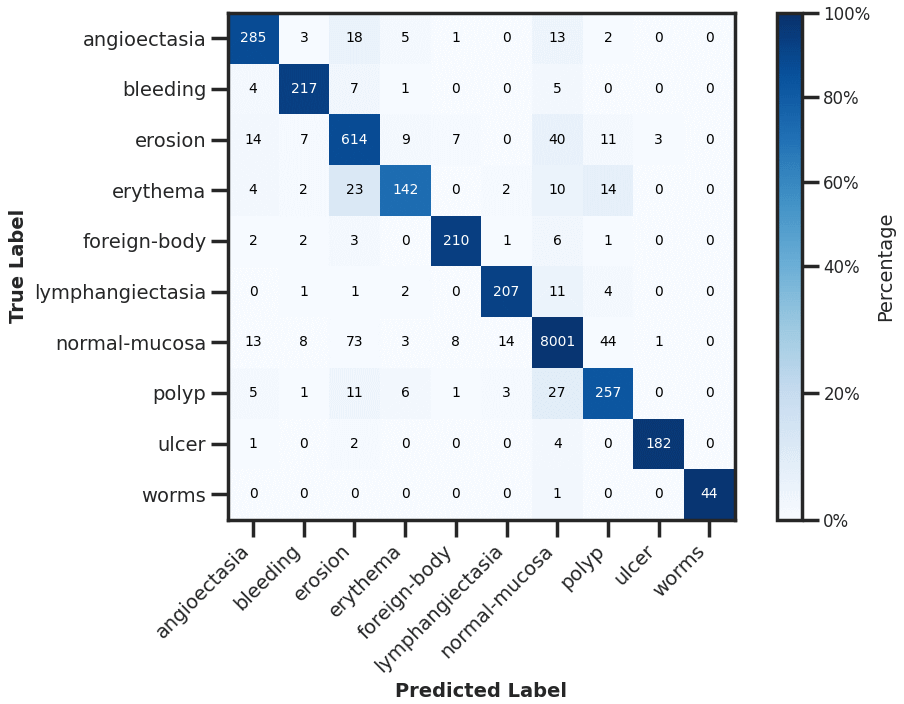}
    \caption{Confusion matrix of our fine-tuned eva02-base on the CE validation set.}
    \label{fig:conf_mat}
\end{figure}

Figure \ref{fig:roc_curves} provides a visual representation of the model's classification performance, as indicated by receiver operating characteristic (ROC) curves. The ROC curve results indicate strong discriminative ability across all gastrointestinal findings. The right panel shows the micro- and macro-average ROC curves, both achieving an AUC of 0.99 and 0.97 respectively, which further underscores the model's robust overall performance in multi-class classification.
The results indicate that the eva02-base model effectively utilizes adaptive pre-training and demonstrates resilience in addressing the distinctive challenges inherent to GIE multi-class classification, as exemplified by the CE24 dataset.
Figure \ref{fig:conf_mat} shows the confusion matrix for the multi-class classification of pathological findings, with the highest accuracy observed for \textit{normal mucosa}, with 8001 true positives, and \textit{worms}, with 44 true positives. Misclassifications are most frequent between similar classes like \textit{erosion} and \textit{normal mucosa}, while rare conditions like \textit{worms} show minimal confusion with other classes.

\section{Conclusion}\label{sec5}
In this work, we presented a novel approach for domain-adaptive pre-training of self-supervised foundation models in the context of GIE diagnostics. By introducing the EndoExtend24 dataset, a comprehensive collection of over 226,000 labeled images across multiple GIE modalities, we addressed key challenges such as dataset scarcity, terminology inconsistency, and data leakage. Our experiments demonstrate the effectiveness of pre-training the EVA-02 model on EndoExtend24 and fine-tuning it on the Capsule Endoscopy 2024 Challenge dataset \cite{Handa2024_trainval}, securing third place in the challenge \cite{handa2024capsule_paper}. We achieved a macro AUC of 0.762 and a balanced accuracy of 37.1\% on the CE24 test set \cite{Handa2024_test}. Notably, we
surpassed the first-place model in balanced accuracy, with our 37.1\% compared to their 35.7\%, despite their higher macro AUC of 0.857 \cite{PuppyOps}. These results highlight the potential of leveraging large-scale, domain-adaptive self-supervised pre-training to enhance the performance of foundation models for specialized downstream tasks such as VCE. Our future work will explore further improvements to dataset standardization and model interpretability to support broader adoption of AI in clinical settings and foster the explainability of model results for enhanced clinical insights.

\section{Acknowledgments}\label{sec6}
As participants in the Capsule Vision 2024 Challenge, we fully comply with the competition's rules as outlined in \cite{handa2024capsule_paper}. Our AI model development is based exclusively on the datasets provided in the official release in \cite{Handa2024_trainval}.

\bibliographystyle{unsrtnat}
\bibliography{main}

\end{document}